\documentclass{article}
\usepackage{amsmath}
\usepackage{amssymb}
\usepackage{multirow}
\usepackage{booktabs}
\usepackage{graphicx}
\usepackage{multicol}
\usepackage{enumitem}
\usepackage[belowskip=-10pt,aboveskip=5pt,font=footnotesize]{caption}
\usepackage{wrapfig}
\usepackage{natbib}

    \usepackage[final]{neurips_2020}

\usepackage[utf8]{inputenc} 
\usepackage[T1]{fontenc}    
\usepackage{hyperref}       
\usepackage{url}            
\usepackage{booktabs}       
\usepackage{amsfonts}       
\usepackage{nicefrac}       
\usepackage{microtype}      

\title{Kronecker Factorization for Preventing Catastrophic Forgetting in Large-scale Medical Entity Linking}

\author{%
  Denis Jered McInerney\thanks{Work completed during an internship at Amazon.}\\
  Northeastern University\\
  \texttt{mcinerney.de@northeastern.edu} \\
   \And
   Luyang Kong \\
   Amazon AI \\
   \texttt{luyankon@amazon.com} \\
   \AND
   Kristjan Arumae \\
   Qualtrics \\
   \texttt{kristjana@qualtrics.com} \\
   \And
   Byron Wallace \\
   Northeastern \\
   \texttt{b.wallace@northeastern.edu} \\
   \And
   Parminder Bhatia \\
   Amazon AI \\
   \texttt{parmib@amazon.com} \\
}

\begin{document}

\maketitle

\begin{abstract}
  
    Multi-task learning is useful in NLP because it is often practically desirable to have a single model that works across a range of tasks.
    In the medical domain, sequential training on tasks may sometimes be the only way to train models, either because access to the original (potentially sensitive) data is no longer available, or simply owing to the computational costs inherent to joint retraining. 
    A major issue inherent to sequential learning, however, is \textit{catastrophic forgetting}, i.e., a substantial drop in accuracy on prior tasks 
    when a model is updated for a new task. 
    Elastic Weight Consolidation is a recently proposed method 
    to address this issue, but scaling this approach to the modern large models used in practice requires making strong independence assumptions about model parameters, limiting its effectiveness.
    In this work, we apply Kronecker Factorization---a recent approach that relaxes independence assumptions---to prevent catastrophic forgetting in convolutional and Transformer-based neural networks at scale.
    We show the effectiveness of this technique 
    on the important and illustrative task of medical entity linking across three datasets, demonstrating the capability of the technique to be used to make efficient updates to existing methods as new medical data becomes available. 
    On average, the proposed method reduces catastrophic forgetting by 51\% when using a BERT-based model, compared to a 27\% reduction using standard Elastic Weight Consolidation, while maintaining spatial complexity proportional to the number of model parameters.
\end{abstract}



\section{Introduction}
\vspace{-0.6em}

Creating a single model that performs well 
across multiple domains is often desirable, especially in production systems. 
Relying on multiple (task-specific) systems necessitates storing and managing corresponding collections of parameters.
\emph{Multi-task models} \cite{caruana1997multitask} obviate this need by performing well on inputs from all tasks, simplifying deployment.

In the medical domain especially, new data is constantly becoming available, and it is necessary to keep models up to date with this deluge. 
In medical entity linking---where the goal is to link mentions in clinical text to corresponding entities in an ontology---the underlying ontologies are frequently updated, and the new terms are put into use quickly. 
For example, in the past year or so many codes related to COVID-19 \cite{doi:10.1056/NEJMoa2002032} were added to the International Classification of Diseases (ICD) lexicon, therefore updating models to incorporate new codes without losing performance on older knowledge is of particular importance. 
The language in the medical domain also brings additional challenges because of the many acronyms, synonyms, and ambiguous terms used in both clinical and biomedical corpora. 
These characteristics make it difficult even for humans to choose the correct entity among the top candidates in an ontology.

To train a multi-task model, one would ideally jointly train on data drawn from all tasks. 
However, when new tasks are introduced, 
this would require re-training on the combined data, which is inefficient and sometimes practically impossible. 
For example, there are cases---particularly when dealing with medical data---where data access is lost, 
precluding joint training over a combined set of old and new data. 
This occurs, e.g., when license agreements or contracts with data providers expire; time-limited data partnerships are common in industry settings.
In these situations, \emph{continuous learning}, or training on task-specific data sequentially, 
is the only option to maintain a single model across tasks. 

Previous work on continuous learning has focused on mitigating \textit{catastrophic forgetting} (CF) \cite{MCCLOSKEY1989109, Ratcliff1990ConnectionistMO}, 
a problem that arises in sequential training where performance on earlier tasks drops when the model is trained on additional tasks.
\emph{Experience Replay} \cite{NIPS2019_9471}
maintains performance when training on new tasks by ``replaying'' examples saved 
from older tasks. 
Unfortunately, in a setting in which access to previous data is 
not possible, this method cannot be used.

Elastic Weight Consolidation (EWC) \cite{Kirkpatrick_2017} is an alternative, constraint-based technique that regularizes parameters such that they are encouraged to maintain optimal weights learned for prior tasks. 
By placing a prior involving the Hessian from previous tasks over network parameters, EWC affords flexibility with respect to changing parameters in different dimensions. 
Critically, EWC does not require continued access to data from `past' tasks once key statistics are computed over a task's data. 
However, to scale to even relatively small neural networks, EWC must assume independence between all parameters. 
This assumption allows one to drop off-diagonal terms in the Fisher Information Matrix (FIM), which is used to approximate the Hessian; calculating the full matrix would be intractable.
Recent work \cite{NIPS2018_7631} has proposed Kronecker Factorization (KF) --- which the optimization community uses to compute Hessians in neural networks --- to perform a version of EWC with a relaxed independence assumption on networks of linear layers operating on small vision datasets.
{\bf Our contribution here is the extension of EWC and KF to the large-scale neural models now common in NLP.} 

In particular, modern NLP tends to rely on large-scale models with hundreds of millions of parameters \cite{Devlin_2019, liu2019roberta, radford2019language}, and CF is problematic across many of its sub-domains.
Though EWC has been successfully applied to NLP in recent work (Section \ref{sec:related_work}), we demonstrate that there is room for substantial gains.
In particular, 
we have observed that the independence assumption over parameters significantly 
and negatively affects EWC's ability to mitigate CF, as compared to what is achieved using the full covariance 
matrix of parameters.

As far as we are aware, this work is the first application of the Kronecker Factorization method---which relaxes the assumption of independence between parameters---for continuous learning in large-scale networks. Though we do not model all elements of the Fisher Information Matrix, we approximate block diagonals corresponding to layers, which is less damaging than assuming completely independent parameters.
Specifically, we apply Kronecker Factorization in two large-scale neural networks for Entity Linking: (1) a convolutional and (2) a transformer-based architecture \cite{NIPS2017_7181}.

Our primary contributions are as follows: (1) We combine and extend prior work in EWC and Kronecker Factorization to modern large-scale NLP models. (2) We use Kronecker Factorization to train on multiple biomedical ontology entity linking tasks sequentially without access to previous data, and show that it significantly outperforms baseline methods.

\section{Continuous Learning Strategies}
\vspace{-.2em}

Here we compare recently proposed baselines, describe an extension of EWC, and demonstrate how to scale this to modern, large NLP models. 
We differentiate between methods that require access to prior task data during training and those that do not.

\textbf{Learning rate control} was proposed in ULMFit 
\cite{howard-ruder-2018-universal} as ``discriminative fine-tuning'' motivated by the intuition that different layers contain distinct information and should be fine-tuned at different rates, accordingly. 
In particular, we control the extent of fine-tuning by altering the learning rate for different layers according to $\eta_\ell = \eta_{\ell-1}\gamma$
where $\eta_\ell$ is the learning rate for layer $\ell$ and $\gamma$ is a constant hyper-parameter. Though the authors apply this method to fine-tuning networks when the pre-training task is not one of the downstream tasks, they suggest that this can also prevent CF, so it serves as a good baseline to our other methods.

\textbf{Experience Replay} 
\cite{NIPS2019_9471}
is a method for preventing CF in continuous learning by ``replaying'' examples from a memory buffer.
After every $m$ training batches on a task, we ``replay'' $n$ batches from prior tasks. 
This requires access to all prior task data while adapting to a new task, whereas our goal is to design an approach that does not require this data.
Still, we compare our methods to experience replay to assess the relative performance we can achieve when we maintain only statistics and model weights for prior tasks.

\textbf{Elastic Weight Consolidation} \cite{Kirkpatrick_2017} mitigates CF using an approximation of a posterior on the optimal weights for prior tasks in order to constrain learning on new tasks:
\begin{equation}
    L_t^{\textrm{EWC}} = L_t - \log p(\theta_t | D_{1:t-1}),
\end{equation}
for task $t$ where $L_t = -\log p(D_t | \theta_t)$ is the loss for that task.
In particular, the posterior is approximated as a Gaussian Mixture Model where each mode corresponds to a prior task. Specifically,
\begin{equation}
    p(\theta_t | D_{1:t-1}) \propto
    \exp{\left(-\lambda \sum_{i=1}^{t-1}(\theta_t - \mu_{i})\Lambda_{i}^{-1}(\theta_t - \mu_{i})\right)}
\end{equation}
where $i$ 
indexes over previous tasks, $\mu_{i}$ and $\Lambda_{i}$ are the mean and covariance for the parameters of task $i$, and $\lambda$ is a hyper-parameter that controls the strength of the regularization or from another view, the variance of the posterior. 
The mean represents the optimal weights for the task, approximated using the parameters directly after training on the task ($\mu_i = \theta^*_i$). 
We estimate the inverse of the covariance matrix with the Fisher Information Matrix (FIM) \cite{frieden_2004, montecarlofisher}
\begin{equation}\label{eq:fisher}
    \Lambda_i^{-1} \approx \mathbb{E}\left[\frac{\partial L_i}{\partial \theta_i} \frac{\partial L_i}{\partial \theta_i}^\intercal\right].
\end{equation}
In practice, for purposes of computational efficiency, we use only the diagonal of the FIM \cite{Kirkpatrick_2017} to compute the regularization term, in turn implying that
\begin{equation}
    p(\theta_t | D_{1:t-1}) \propto 
    \exp{\left(-\lambda \sum_{i=1}^{t-1}\sum_j\Lambda_{i,(j,j)}^{-1}(\theta_{t,j} - \mu_{i,j})^2\right)}
\end{equation}
where $j$ indexes over the parameters.

Unlike experience replay, this method has the benefit of not requiring access to prior task data during training, provided that the FIM is computed ahead of time. 
This also holds for our main method, Kronecker Factorization. 
Though EWC and Kronecker Factorization both require keeping a set a weights and a FIM from after training on each task, these can be discarded when performing inference. 

\begin{figure*}
    \centering
    \includegraphics[scale=.75]{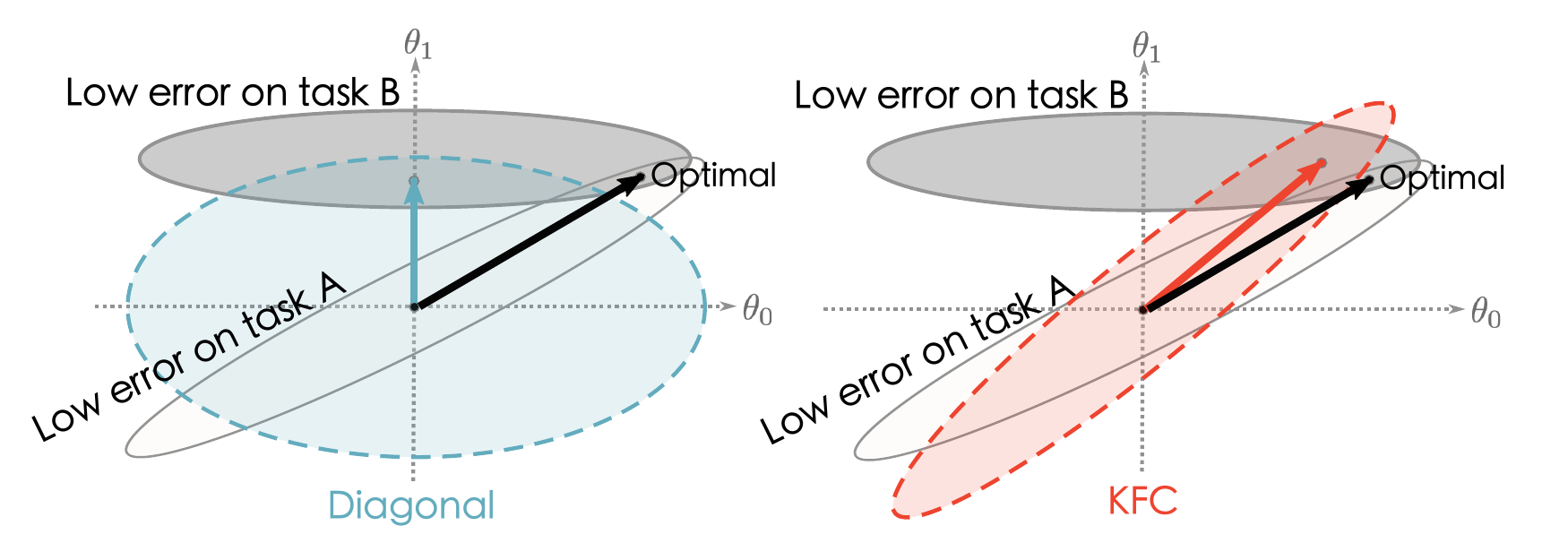}
    \caption{\textbf{Toy 2-parameter example}. The diagonal approximation of the Fisher Information Matrix (FIM) assumes that $\theta_0$ and $\theta_1$ are independent, so the corresponding Gaussian ellipse has no tilt. The Kronecker-Factored approximation to the FIM may not exactly align with the true local Gaussian over the optimal parameters for task A, but it can still model interactions between parameters, so it has a similar tilt to the true Gaussian. Using this KFC approximation, the estimated parameters are much closer to the optimal than when using the diagonal FIM.}
    \label{fig:kfac_concept}
\end{figure*}

\textbf{Kronecker Factorization.} A downside to using a diagonal FIM is that this assumes independence between all model parameters. 
However, $\Lambda_i^{-1}$ is an $n \times n$ matrix where $n$ is the number of parameters in the model, so computing this for contemporary neural networks with millions of parameters is computationally infeasible. 
\citet{NIPS2018_7631} relax this independence assumption while still maintaining low memory requirements by using a convenient Kronecker Factorization of the FIM for parameters within a layer.

For a linear layer without bias $W \in \mathbb{R}^{p \times q}$, input $\boldsymbol{x} \in \mathbb{R}^{p \times 1}$, and output $\boldsymbol{y} = W^\intercal \boldsymbol{x}$ where $\boldsymbol{y} \in \mathbb{R}^{q \times 1}$, the derivative of the weights can be written as $\frac{\partial L}{\partial W} = \boldsymbol{x} \frac{\partial L}{\partial \boldsymbol{y}}^\intercal$.
We elide indexing over the task and the instance for clarity. Therefore, using Equation \ref{eq:fisher}, the inverse of the full covariance can be factored as
\begin{equation}
    \mathbb{E}\left[\mathrm{vec}(\boldsymbol{x} \frac{\partial L}{\partial \boldsymbol{y}}^\intercal)\mathrm{vec}(\boldsymbol{x} \frac{\partial L}{\partial \boldsymbol{y}}^\intercal)^\intercal\right] =
    \mathbb{E}\left[\boldsymbol{x}\boldsymbol{x}^\intercal \otimes \frac{\partial L}{\partial \boldsymbol{y}}\frac{\partial L}{\partial \boldsymbol{y}}^\intercal\right].
\end{equation}
Then, making an assumption of independence between the two Kronecker factors $\boldsymbol{x}\boldsymbol{x}^\intercal$ and $\frac{\partial L}{\partial \boldsymbol{y}}\frac{\partial L}{\partial \boldsymbol{y}}^\intercal$, we can approximate the full covariance for the linear layer as
\begin{equation}\label{eq:kfac_fisher_approx}
    \mathbb{E}\left[\boldsymbol{x}\boldsymbol{x}^\intercal \otimes \frac{\partial L}{\partial \boldsymbol{y}}\frac{\partial L}{\partial \boldsymbol{y}}^\intercal\right] \approx \mathbb{E}\left[\boldsymbol{x}\boldsymbol{x}^\intercal\right] \otimes \mathbb{E}\left[\frac{\partial L}{\partial \boldsymbol{y}}\frac{\partial L}{\partial \boldsymbol{y}}^\intercal\right].
\end{equation}
We can extend this to layers with bias by appending a constant to the input and adding a column to the weights: $\boldsymbol{x}' = \mathrm{Concat}(\boldsymbol{x}, [1])$ and $W' = \mathrm{Concat}(W, \boldsymbol{b}^\intercal)$ where $\boldsymbol{b} \in \mathbb{R}^{q \times 1}$.

Figure \ref{fig:kfac_concept} illustrates the motivation for including off-diagonal terms in the FIM. 
In particular, in the optimal parameter zones for tasks A and B in the figure, estimating off-diagonal terms is beneficial in achieving the near-optimal parameters for both tasks.

\textbf{Spatial Complexity.} 
The approximation to the full FIM in Equation \ref{eq:kfac_fisher_approx} allows us to store only the factors in place of the full FIM. The spatial complexity of the full FIM is $P^2$ where $P$ is the number of parameters in the network.  
In general, neural networks have a structure that makes parameters have $O(d^2 l)$ complexity where the intermediate dimensions are all $O(d)$ and $l$ is the number of layers.  
This means that the number of elements in the full FIM is $O(d^4 l^2)$.  
Assuming layer-wise independence makes this a block-diagonal FIM which has $O(d^4 l)$ non-zero elements.  
Finally, because Kronecker Factorization reduces the memory to $p^2 + q^2$ elements instead of $p^2 q^2$ for an individual linear layer with input dimension $p$ and output dimension $q$, the full memory usage is $O(d^2 l)$. 
Given a slightly stronger assumption that $P = \Theta(d^2l)$ instead of just $O(d^2l)$, which holds for the networks in this paper and most others, we see that this method, like EWC, is also linearly dependent on the parameters.

We demonstrate the complexity of full FIM complexity by reducing hidden size until a model becomes practically feasible to train. 
We reduced the hidden dimensionality of our smaller convolutional model from $768$ to $10$ in order to compute the full FIM.  
An increase of this dimensionality to $20$ makes the model too large to be held in memory for both the full FIM as well as the layer-wise independent block-diagonal FIM.  
Kronecker Factorization is therefore essential to reduce asymptotic memory requirements. 

\textbf{Scaling to Large Models.} We use two models to perform entity linking: 1) a CNN-based model and 2) a Transformer-based model.
In order to implement Kronecker Factorization on the CNN, we must extend it to convolutional layers using the derivations in \cite{pmlr-v48-grosse16}.
A CNN can be thought of as a variation of a linear layer in which the linear layer is applied to multiple inputs in one forward pass of the model.
In this case, the expectation is not only over each instance in the dataset, but also over each time the linear layer is used with different inputs.
In order to take this expectation, we make the assumption that the gradients with respect to these inputs are independent even within a single instance.
We also make this assumption for models with shared weights in different linear layers. 
Since the only layers in Transformer models are linear layers and layer norms, we decide to ignore layer norms in our EWC regularization term. Though we do not include a recurrent neural network (RNN) model in this paper, these methods can be easily extended to such architectures.\footnote{
RNNs are constructed of linear layers with non-linear activations. 
All that is needed is to assume independence between the gradients with respect to different inputs to shared parameters and average the factors calculated from these different inputs paired with their outputs' gradients. 
In the case of RNNs, these different inputs are different time-steps.}



\section{Datasets}



\begin{table}[t]
    \parbox{.47\linewidth}{
    \footnotesize
    \centering
    \begin{tabular}{l c| r r}
        \multirow{2}{*}{Perm.} & \multirow{2}{*}{Method} & \multicolumn{2}{c}{MedM.} \\
        & & Acc. & Perc. $\Delta$ \\
        \hline
        MedM. & - & 74.85 & - \\
        \hline
        \multirow{5}{*}{MedM. $\rightarrow$ 3DNotes} & None & 66.27 & -11.46 \\
        & LRC & 69.34 & -7.36 \\
        & ER & 73.45 & -1.87 \\
        & L2 Norm & 68.45 & -8.55 \\
        & EWC & 72.17 & -3.58 \\
    \end{tabular}
    \caption{\textbf{Comparison of baseline catastrophic forgetting mitigation methods} on subsets on down-sampled data (comprising 10\% of instances).}
    \label{tab:preliminary_results}
    }
    \hfill
    \parbox{.47\linewidth}{
    \footnotesize
    \centering
    \begin{tabular}{l l|r r}
        Training & Dataset & CNN & BERT \\
        \hline
        \multirow{3}{*}{Individual} & MedM. & 76.75 & 84.82 \\
        & 3DNotes & 77.89 & 80.16 \\
        & MedN. & 88.47 & 90.95 \\
        \hline
        \multirow{3}{*}{Combined} & MedM. & 76.26 & 81.88 \\
        & 3DNotes & 78.34 & 79.24 \\
        & MedN. & 89.55 & 92.46 \\
    \end{tabular}
    \caption{\textbf{Individual and Combined Training.} Entity linking accuracy after training individual models for each dataset on full data and one combined model on all datasets shuffled together.}
    \label{tab:tier_1}
    }
\end{table}

Entity linking is a longstanding task in NLP that entails matching the mentions, spans of text in natural language, with the entities to which they refer in an ontology. In this paper, we assume that each instance consists of a trimmed down set of candidates, and a mention in its context. 
Specific to our motivating setting of bio-medical text, there exist a large set of diverse ontologies in the medical domain, and it is common to want to link entities in clinical texts to these.
In such applications it is not uncommon to lose access to (sensitive) training datasets; but in many cases we would like to maintain model performance on the tasks that these represent. 
We use all possible permutations of the Medmention, MedNorm, and 3DNotes datasets to test how well different methods prevent CF. For details on the statistics, we refer to the appendix.

{\bf MedMention} \cite{medmention_dataset} is a publicly available corpus for medical concept normalization. 
It conrains 4,392 abstracts from PubMed\footnote{https://www.ncbi.nlm.nih.gov/pmc/}, with bio-medical entities annotated with Unified Medical Language System (UMLS) concepts. 
We follow the approach of candidate generation described in \citet{murty-hierarchical}. 
We retain only the top nine most similar entities (excluding the ground truth entity) as negative candidates. 
In addition, the ground truth entity will be considered as the positive candidate, thus forming a set of 10 candidates for each mention.

{\bf MedNorm} \cite{mednorm-dataset} is a corpus of 27,979 
descriptions 
mapped to two medical ontologies 
(MedDRA and SNOMED-CT), sourced from five publicly available datasets across
biomedical and social media domains. 
We use a subset of these: CADEC \cite{KARIMI201573}, TAC \cite{demner2018dataset}, and TwiMed \cite{alvaro2017twimed} and employ a BM25 \cite{Robertson1994SomeSE} based retrieval approach to generate top entities for the mention, which will be candidates of the ranking model.

{\bf 3DNotes}. We also use a de-identified corpus of dictated doctor’s notes (3DNotes), similar to \citet{LATTE}. 
These are annotated with medical entities related to signs, symptoms, and diseases. These entities are mapped to the 10th version of International Statistical Classification of Diseases and related health problems (ICD-10), which is part of UMLS. 
The annotation guidelines are similar to the i2b2 challenge \cite{i2b2} guidelines for the problem entity. 
We use 20 words on the left and right as mention context.

\section{Experiments}\label{sec:experiments}

We perform initial experiements using a downsampled dataset (comprising ten percent of the data) and one ordering over two datasets (MedMention, 3DNotes) to compare Learning Rate Control, Experience Replay, L$_2$ regularization, and EWC. 
Of these, we take EWC as the baseline from which to compare the proposed KFC extension.\footnote{We do not perform further explicit comparisons to Experience Replay, as we are interested in approaches that do not assume access to data from all tasks.} 
\begin{table*}[t]
    \footnotesize
    \centering
    \begin{tabular}{c c||r r r|r r r}
        \multirow{3}{*}{Perm.} & \multirow{3}{*}{Reg.} & \multicolumn{3}{c|}{CNN} & \multicolumn{3}{c}{BERT} \\
        & & \multicolumn{2}{c|}{Task A} & Task B & \multicolumn{2}{c|}{Task A} & Task B \\
        & & \multicolumn{1}{c|}{Acc.} & \multicolumn{1}{c|}{Perc. $\Delta$} & Acc. & \multicolumn{1}{c|}{Acc.} & \multicolumn{1}{c|}{Perc. $\Delta$} & Acc. \\
        \toprule
        
        \multirow{3}{*}{MedM. $\rightarrow$ 3DNotes} & None & 41.35 & -46.12 & \textbf{77.54} & 63.77 & -24.82 & 79.48 \\
        & EWC (diag.) & 55.41 & -27.81 & 75.81 & 76.41 & -9.91 & 79.01 \\
        & Ours (KFC) & \textbf{68.55} & \textbf{-10.68} & 74.05 & \textbf{80.64} & \textbf{-4.93} & \textbf{79.78} \\
        \hline
	
        \multirow{3}{*}{MedM. $\rightarrow$ MedN.} & None & 48.55 & -36.75 & 89.75 & 60.53 & -28.64 & 90.75 \\
        & EWC (diag.) & 56.98 & -25.75 & 90.07 & 70.04 & -17.42 & \textbf{91.21} \\
        & Ours (KFC) & \textbf{73.93} & \textbf{-3.67} & \textbf{90.09} & \textbf{76.64} & \textbf{-9.64} & 90.35 \\
        \hline
	
        \multirow{3}{*}{3DNotes $\rightarrow$ MedM.} & None & 46.48 & -40.32 & \textbf{77.04} & 55.36 & -30.95 & \textbf{83.43} \\
        & EWC (diag.) & 55.29 & -29.02 & 76.67 & 61.26 & -23.59 & 82.32 \\
        & Ours (KFC) & \textbf{66.18} & \textbf{-15.04} & 74.13 & \textbf{71.81} & \textbf{-10.42} & 83.15 \\
        \hline
	
        \multirow{3}{*}{3DNotes $\rightarrow$ MedN.} & None & 47.13 & -39.49 & \textbf{90.78} & 35.89 & -55.23 & \textbf{90.89} \\
        & EWC (diag.) & 58.42 & -24.99 & 86.38 & 48.95 & -38.94 & 90.61 \\
        & Ours (KFC) & \textbf{72.13} & \textbf{-7.39} & 88.95 & \textbf{75.04} & \textbf{-6.40} & 89.27 \\
        \hline
	
        \multirow{3}{*}{MedN. $\rightarrow$ MedM.} & None & 49.59 & -43.95 & \textbf{75.93} & 25.38 & -72.10 & \textbf{84.20} \\
        & EWC (diag.) & 67.74 & -23.43 & 73.20 & 22.78 & -74.95 & 83.34 \\
        & Ours (KFC) & \textbf{74.59} & \textbf{-15.68} & 74.39 & \textbf{25.41} & \textbf{-72.07} & 82.77 \\
        \hline
	
        \multirow{3}{*}{MedN. $\rightarrow$ 3DNotes} & None & 55.81 & -36.92 & \textbf{76.73} & 19.01 & -79.10 & \textbf{80.34} \\
        & EWC (diag.) & 64.60 & -26.98 & 75.89 & \textbf{29.29} & \textbf{-67.80} & 79.43 \\
        & Ours (KFC) & \textbf{79.22} & \textbf{-10.45} & 74.68 & 24.89 & -72.63 & 77.28 \\
        \hline
	
    \end{tabular}
    \caption{\textbf{Tier 2 Results.} Entity Linking accuracy on Task A and B after further training the tier 1 models (trained on task A, the first dataset in the permutation) on task B, the second dataset in the permutation. The percentage change represents the difference in performance between the original tier 1 model and the tier 2 model.}
    \label{tab:tier_2}
\end{table*}

\begin{table*}[t]
    \centering
    \resizebox{\textwidth}{!}{
    \begin{tabular}{c c||r r r r r|r r r r r}
        \multirow{3}{*}{Perm.} & \multirow{3}{*}{Reg.} & \multicolumn{5}{c|}{CNN} & \multicolumn{5}{c}{BERT} \\
        & & \multicolumn{2}{c|}{Task A} & \multicolumn{2}{c|}{Task B} & Task C & \multicolumn{2}{c|}{Task A} & \multicolumn{2}{c|}{Task B} & Task C \\
        & & \multicolumn{1}{c|}{Acc.} & \multicolumn{1}{c|}{Perc. $\Delta$} & \multicolumn{1}{c|}{Acc.} & \multicolumn{1}{c|}{Perc. $\Delta$} & Acc. & \multicolumn{1}{c|}{Acc.} & \multicolumn{1}{c|}{Perc. $\Delta$} & \multicolumn{1}{c|}{Acc.} & \multicolumn{1}{c|}{Perc. $\Delta$} & Acc. \\
        \toprule
        
        MedM. & None & 25.09 & -67.31 & 44.50 & -42.86 & 90.52 & 34.31 & -59.54 & 32.28 & -59.73 & \textbf{90.89} \\
        $\rightarrow$ 3DNotes & EWC (diag.) & 52.54 & -31.55 & 29.57 & -62.04 & 90.75 & 48.74 & -42.53 & 36.01 & -55.08 & 88.61 \\
        $\rightarrow$ MedN. & Ours (KFC) & \textbf{72.79} & \textbf{-5.16} & \textbf{48.70} & \textbf{-37.47} & \textbf{91.64} & \textbf{62.05} & \textbf{-26.85} & \textbf{57.06} & \textbf{-28.83} & 89.38 \\
        \hline
	
        MedM. & None & 37.38 & -51.30 & 42.73 & -51.69 & \textbf{77.19} & 58.70 & -30.79 & 47.05 & -48.27 & \textbf{80.71} \\
        $\rightarrow$ MedN. & EWC (diag.) & \textbf{65.67} & \textbf{-14.43} & 70.40 & -20.43 & 58.19 & 76.25 & -10.10 & 45.45 & -50.03 & 79.57 \\
        $\rightarrow$ 3DNotes & Ours (KFC) & 65.28 & -14.94 & \textbf{81.42} & \textbf{-7.97} & 66.53 & \textbf{76.81} & \textbf{-9.45} & \textbf{48.44} & \textbf{-46.74} & 75.82 \\
        \hline
	
        3DNotes & None & 23.97 & -69.23 & 45.25 & -41.04 & \textbf{90.69} & 16.14 & -79.86 & 59.43 & -29.93 & \textbf{91.44} \\
        $\rightarrow$ MedM. & EWC (diag.) & 39.74 & -48.98 & 48.10 & -37.32 & 90.21 & 22.78 & -71.59 & 46.40 & -45.29 & 88.58 \\
        $\rightarrow$ MedN. & Ours (KFC) & \textbf{64.27} & \textbf{-17.49} & \textbf{70.43} & \textbf{-8.23} & 89.67 & \textbf{48.34} & \textbf{-39.70} & \textbf{78.15} & \textbf{-7.86} & 89.78 \\
        \hline
	
        3DNotes & None & 43.19 & -44.55 & 55.38 & -37.40 & \textbf{76.72} & 53.78 & -32.91 & \textbf{35.26} & \textbf{-61.24} & \textbf{83.50} \\
        $\rightarrow$ MedN. & EWC (diag.) & 57.98 & -25.56 & 51.04 & -42.30 & 62.94 & \textbf{53.97} & \textbf{-32.67} & 29.55 & -67.51 & 82.28 \\
        $\rightarrow$ MedM. & Ours (KFC) & \textbf{64.60} & \textbf{-17.06} & \textbf{80.56} & \textbf{-8.94} & 62.82 & 47.08 & -41.28 & 27.92 & -69.30 & 79.13 \\
        \hline
	
        MedN. & None & 24.55 & -72.25 & 38.83 & -49.40 & \textbf{76.49} & \textbf{27.46} & \textbf{-69.81} & 62.51 & -26.30 & \textbf{80.81} \\
        $\rightarrow$ MedM. & EWC (diag.) & 71.08 & -19.65 & 53.46 & -30.35 & 59.30 & 24.21 & -73.38 & 56.92 & -32.90 & 79.80 \\
        $\rightarrow$ 3DNotes & Ours (KFC) & \textbf{72.65} & \textbf{-17.88} & \textbf{66.26} & \textbf{-13.66} & 71.90 & 23.38 & -74.29 & \textbf{71.98} & \textbf{-15.14} & 79.17 \\
        \hline
	
        MedN. & None & 39.57 & -55.28 & 47.99 & -38.39 & \textbf{76.40} & \textbf{26.83} & \textbf{-70.50} & 52.47 & -34.55 & \textbf{83.62} \\
        $\rightarrow$ 3DNotes & EWC (diag.) & 48.73 & -44.92 & 64.72 & -16.90 & 64.51 & 16.93 & -81.39 & 43.93 & -45.21 & 82.54 \\
        $\rightarrow$ MedM. & Ours (KFC) & \textbf{73.74} & \textbf{-16.65} & \textbf{71.48} & \textbf{-8.23} & 64.07 & 21.38 & -76.49 & \textbf{64.93} & \textbf{-19.00} & 82.74 \\
        \hline

    \end{tabular}
    }
    \caption{\textbf{Tier 3 Results.} Similar to table \ref{tab:tier_2}, but after further training the tier 2 models on a 3rd dataset.}
    \label{tab:tier_3}
\end{table*}
Specifically, we compare traditional EWC (using the diagonal FIM) to one using Kronecker Factorization on the full data and all possible permutations of all three datasets. 
We select optimal $\lambda$s using the development sets in each dataset, and report results on the test set for those $\lambda$s. 
In Section \ref{app:hyperparameter_search} of the appendix, we further discuss the hyper-parameter search which can give a sense of the trade-off between low and high regularization coefficients. We perform all experiments using a CNN-based model with 26M parameters and a Transformer-based model with 46M parameters that is initialized using the first three layers of BERT \cite{Devlin_2019}, training for 10 epochs on each dataset. We perform each tier of training for models using an NVIDIA V100 GPU with 16GB of memory.

\section{Results}


Our initial comparison of baseline methods using a subset of the full data (Table \ref{tab:preliminary_results}) shows that of the methods that do not require access to data from prior tasks (i.e., all save for Experience Replay), EWC performs the best. 
In all tables, entity linking accuracy is defined as the percentage of mentions where the correct candidate is ranked the highest. 

We then evaluate baseline individual models for each of the datasets shown in Table \ref{tab:tier_1}. 
The BERT model has the highest accuracy across the board. 
This table also shows that the MedNorm dataset is the easiest data on which to perform well. 
For comparison, we also train our models on the combined data of all three datasets and show our results in Table \ref{tab:tier_1}. 
We find similar performance to the individual training, which means that there does exist an almost optimal set of parameters that performs well across all three datasets. 
Generally speaking we expect and verify that training on tasks sequentially will decrease performance on prior tasks due to CF, and we measure this drop by calculating the difference in accuracy between our individually trained models in Table \ref{tab:tier_1} and those after training on a second and third task, shown in the percentage change columns of Tables \ref{tab:tier_2} and \ref{tab:tier_3}.

Tables \ref{tab:tier_2} and \ref{tab:tier_3} show the robustness of Kronecker Factorization in preventing CF on prior tasks. 
In almost all permutations of the datasets for both CNN and BERT models, using the Kronecker Factorization of the FIM outperforms standard EWC. 
These results also show that neither method prevents the model from adapting to new tasks, a potential problem called intransigence \cite{Chaudhry_2018} common for constraint-based methods. We avoid this problem by tuning the regularization strength $\lambda$ (Section \ref{app:hyperparameter_search}). 

In order to visualize the benefits of estimating a FIM with non-zero off-diagonal terms and the degree to which the Kronecker Factorization can accurately estimate the Full Fisher in practice, we present Figure \ref{fig:fisher}.  This shows that there are indeed large off-diagonal terms in the full FIM and some of those are estimated accurately by the Kronecker-Factored FIM.

\vspace{-1em}
\section{Discussion}

\begin{wrapfigure}{R}{0.5\textwidth}
    \vspace{-1em}
    \centering
    \includegraphics[scale=.16]{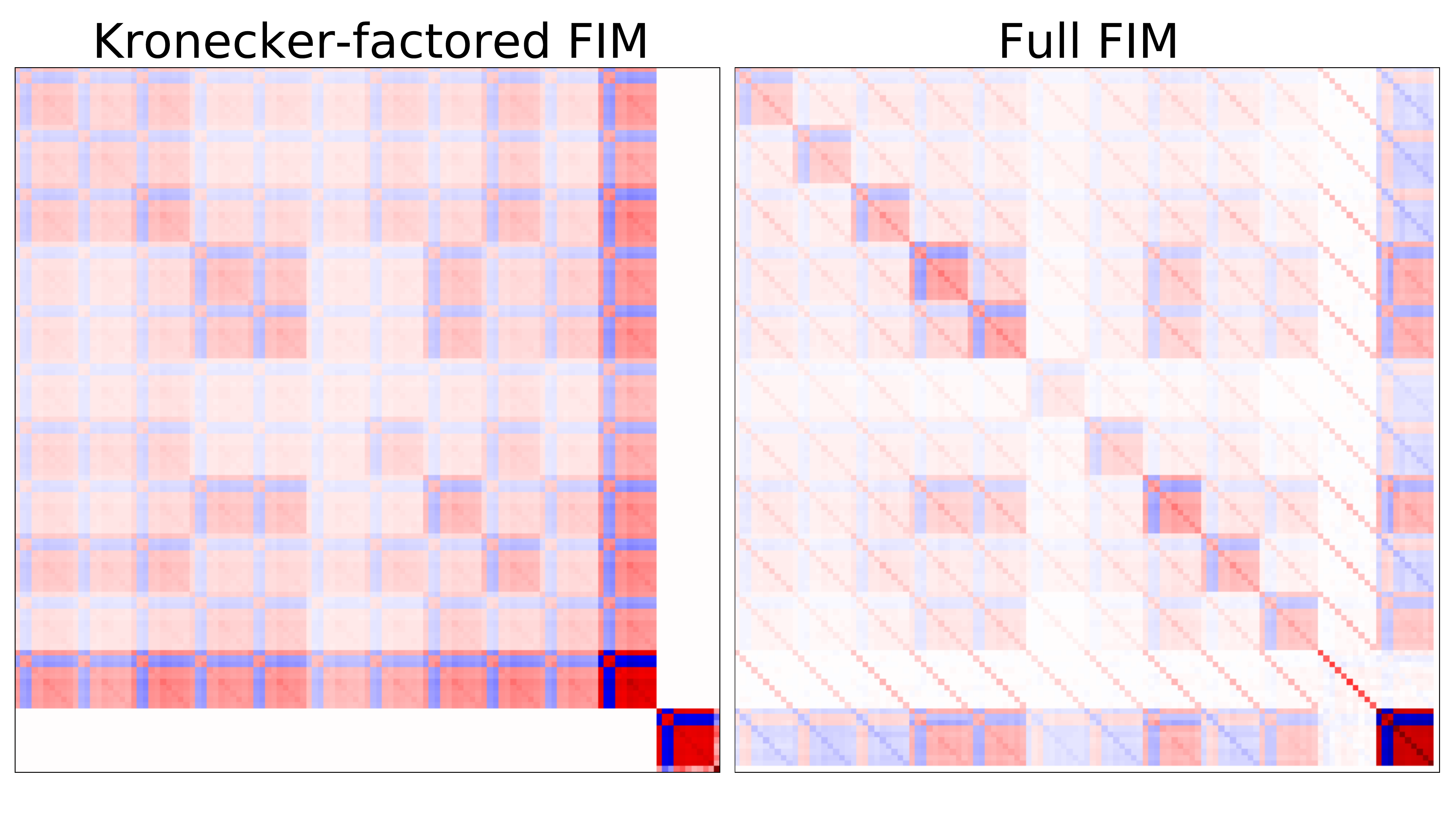}
    \caption{\textbf{Fisher Information Matrix comparison} for a subset of layers in a toy model.}
    \label{fig:fisher}
\end{wrapfigure}

Overall, EWC mitigates CF when compared with the baseline of no regularization, and KFC performs better than EWC, without much hindrance of adaptation to the new task. 
However, there are cases where one or both methods perform worse than the baseline at CF mitigation. 
For instance, in BERT training for permutation MedNorm $\rightarrow$ MedMention, EWC actually does worse than the baseline, and KFC performs about equivalently to the baseline.  
In the BERT permutation 3DNotes $\rightarrow$ MedNorm $\rightarrow$ MedMention. 
On 3DNotes (Task A), KFC performs worse than the baseline and EWC performs equivalently to it. 
In both of the tier 3 training permutations, both regularization strategies perform worse than the benchline on Task A.

These problems appear to occur mostly with BERT (and not with CNNs), and occur when MedNorm is the first task. 
MedNorm is the smallest dataset, so we expect CF to be the worst for it because the number of training steps taken on it is small compared to that which is taken on the other datasets after it. 
The fact that these problems occur mostly with BERT may be due to the larger number of parameters or more complex parameter inter-dependence resulting in a harder-to-approximate FIM, however, this is only a conjecture at present.

We observe that in some cases, performance on the new task is completely unchanged or slightly improved. 
This occurs mostly when the new task is MedNorm because the comparative size makes it easier to adapt faster to the new data regardless of the regularization. In other words, the amount of parameter shift needed for this dataset is minimal.

Taking a closer look at Figure \ref{fig:fisher} we can get a better idea of how Kronecker Factorization may be helping. 
The Kronecker-Factorized block-diagonal FIM shows the last two layers of the CNN architecture: 1) a $10 \times 10$ linear layer, and 2) a $10 \times 1$ linear layer.  
The first layer consists of $11 \times 11$ small blocks, some of which share commonalities. The first $10 \times 10$ blocks represent the interactions of one column of the weight matrix with another or itself, and the last row and column represent the interaction of the bias vector with each of the columns of the weight matrix and itself. Though there are some major differences between the Kronecker-Factored and Full FIMs, there is a pattern of similar small blocks that appear in both. In addition, the 6th and 7th rows and columns of small blocks have noticeably smaller magnitudes in both the Kronecker-Factored and Full FIMs.

Figure \ref{fig:fisher} also reveals potential pitfalls of Kronecker Factorization. 
In the Full FIM, one of the more interesting features of each of these small blocks representing column-to-column interactions within the weight matrix is that they tend to have strong diagonals, meaning that elements in the same row of the weight matrix interact strongly. In addition, the blocks on the diagonal representing within-column interaction have elevated intensity, indicating that elements in the same column of the weight matrix also interact strongly. Neither of these phenomena appear in the Kronecker-Factored approximation. 
This provides clues into how we might further sparsify the intra-layer interactions, which may present an alternative or an improvement to Kronecker Factorization. 

Another feature of Figure \ref{fig:fisher} is the high magnitude of elements representing inter-layer interactions which are not modeled in the Kronecker-Factored FIM. This shows that there is still a significant gap between this approximation and using the full Fisher.


\section{Related Work}\label{sec:related_work}

\textbf{CF Mitigation in Continuous Learning.} There is a large body of work on CF mitigation, much of which focuses on constraining model parameters during training. From this category, we use learning rate control from UMLFit \cite{howard-ruder-2018-universal} and EWC \cite{Kirkpatrick_2017} as baselines. 
Path Integral \cite{pmlr-v70-zenke17a} is similar to EWC but calculates weight variances from the optimization path instead of at the end of training. 
\citet{Chaudhry_2018} combines these methods and generalizes the fisher by replacing it with a KL divergence between output distributions of the previous task weights and those of the current weights.
Progressive networks \cite{rusu2016progressive} do not constrain training but 
use features of previously trained networks as additional input to subsequently trained ones. 
Progress and Compress \cite{pmlr-v80-schwarz18a} extends this by using EWC to create one model that performs well on all tasks.

Previous work has also focused on preventing CF by introducing a working memory that allows past examples to be replayed. We use experience replay from \citet{NIPS2019_9471} as an additional data-dependent baseline for our methods, but there are many other methods in this category \cite{sprechmann2018memorybased, wang-etal-2019-sentence}, and many which combine these methods with regularization methods \cite{NIPS2017_f8752278, chaudhry2018efficient}.  

\textbf{Kronecker Factorization for Neural Networks.} \citet{pmlr-v37-martens15} first introduce Kronecker Factorization as an approximation of blocks of the FIM of neural networks. They use this to perform second-order optimization techniques on linear neural networks, and then extend this to optimizing convolutional architectures \cite{pmlr-v48-grosse16}. More recently, \cite{Ritter2018ASL} show that the Fisher approximation can be used as a posterior on network weights, and then expand EWC to use off-diagonal elements of the FIM in its regularization term with this approximation \cite{NIPS2018_7631}. This last paper mainly focuses on small vision datasets.

\textbf{Continuous Learning in NLP.} NLP seems to be particularly susceptible to CF \cite{howard-ruder-2018-universal,Yogatama2019LearningAE}. 
Recent work has therefore focused on developing continuous learning techniques in NLP to mitigate this issue \cite{Moeed2020AnEO, Pilault2020ConditionallyAM, chen-etal-2020-recall}.
There has also been work that applies previous techniques to specific domains
(e.g. machine translation \cite{thompson-etal-2019-overcoming}, sentiment analysis \cite{madasu2020sequential}, and reading comprehension \cite{xu2020forget}).

Many of these methods focus on avoiding over-fitting to \textit{new} tasks during fine-tuning, whereas we focus on maintaining high performance on \textit{old} tasks. 
None of these use Kronecker Factorization, which has not yet been scaled to prevent CF in large NLP models.

CF mitigation is particularly important in clinical NLP given that many clinical datasets are quite different from generic domains and from each other. 
\citet{Arumae2020AnEI} explore CF in language modeling when transferring between the generic, clinical, and biomedical domains and compare learning rate control, experience replay, and EWC.

\section{Conclusions}

We have demonstrated the effectiveness of Kroneker Factorization (KFC) for preventing catastrophic forgetting in modern large-scale neural architectures commonly used in NLP, improving on the results of Elastic Weight Consolidation.
We showed that KFC can be used to create a unified model on multiple domains of Medical Entity Linking with good performance across tasks after a continuous (sequential) learning process.
We highlighted strengths and weaknesses of the adopted approximation used in KFC, pointing to potential future directions.

Future work might consider alternatives to the block diagonal structure on the covariance matrix used in KFC. 
Another promising line of inquiry concerns reducing the difficulty in selecting the $\lambda$ hyper-parameter, which controls the strength of the regularization, without requiring a robust grid-search.







\bibliography{sample}
\bibliographystyle{plainnat}

\appendix

\clearpage

\begin{table}[t]
    \centering
    \begin{tabular}{l|r r r}
         & MedM. &  3DNotes & MedN. \\
         & (Task A) & (Task B) & (Task C) \\
         \hline
        Train Queries & 109,169 & 54,813 & 9,544 \\
        Dev Queries & 35,752 & 7,310 & 1,255 \\
        Test Queries & 35,974 & 5,712 & 4,192 \\
        Cand./Query & 10 & 25 & 10 \\
        Context & 10 & 20 & 10 \\
    \end{tabular}
    \caption{\textbf{Dataset Statistics.}}
    \label{tab:datasets}
\end{table}

\section{Hyperparameter Search}\label{app:hyperparameter_search}

\begin{table*}[t]
    \centering
    \begin{tabular}{c c||c c|c c}
        \multirow{2}{*}{Perm.} & \multirow{2}{*}{Reg.} & \multicolumn{2}{c|}{CNN} & \multicolumn{2}{c}{BERT} \\
        & & \multicolumn{1}{c|}{Task A} & \multicolumn{1}{c|}{Task B} & \multicolumn{1}{c|}{Task A} & \multicolumn{1}{c}{Task B} \\
        \hline
        
        \multirow{2}{*}{MedM. $\rightarrow$ 3DNotes} & EWC (diag.) & 1e3 & 1e2 & 1e5 & 1e5\\
        & Ours (KFC) & 1e3 & 1e1 & 1e3 & 1e2\\
        \hline
        
        \multirow{2}{*}{MedM. $\rightarrow$ MedN.} & EWC (diag.) & 1e3 & 1e6 & 1e5 & 1e5\\
        & Ours (KFC) & 1e3 & 1e5 & 1e3 & 1e5\\
        \hline
        
        \multirow{2}{*}{3DNotes $\rightarrow$ MedM.} & EWC (diag.) & 1e3 & 1e3 & 1e5 & 1e5\\
        & Ours (KFC) & 1e3 & 1e3 & 1e1 & 1e3\\
        \hline
        
        \multirow{2}{*}{3DNotes $\rightarrow$ MedN.} & EWC (diag.) & 1e5 & 1e4 & 1e5 & 1e5\\
        & Ours (KFC) & 1e3 & 1e7 & 1e3 & 1e5\\
        \hline
        
        \multirow{2}{*}{MedN. $\rightarrow$ MedM.} & EWC (diag.) & 1e7 & 1e3 & 1e7 & 1e3\\
        & Ours (KFC) & 1e5 & 1e3 & 1e7 & 1e1\\
        \hline
        
        \multirow{2}{*}{MedN. $\rightarrow$ 3DNotes} & EWC (diag.) & 1e5 & 1e4 & 1e7 & 1e5\\
        & Ours (KFC) & 1e5 & 1e3 & 1e7 & 1e2\\
        \hline
        
    \end{tabular}
    \caption{\textbf{Optimal $\lambda$ Hyperparameter.}}
    \label{tab:optimal_lambdas}
\end{table*}

The early experiments on the smaller amounts of data (see section \ref{sec:experiments}) gave us a sense of the magnitude of optimal lambdas, so for the 2nd tier of training on the full data, we perform a grid search at the following lambdas for both the original diagonal EWC and the Kronecker Factored EWC: 1e1, 1e3, 1e5, 1e7, 1e9.  The plots in the appendix demonstrate that the lowest of these lambdas correspond to almost no CF mitigation compared with no regularization and the highest of these correspond to almost no CF but very poor performance on the 2nd dataset.  This means that these lambdas span the space well. For each of the 6 dataset permutations at the 2nd tier training, we pick the optimal lambda and further use this optimal lambda in the 3rd tier of training.

In the 3rd tier of training, there are actually two hyper-parameters to pick for the regularization terms corresponding to each of the first two datasets. Because the first dataset had a lambda tuned for it during the 2nd tier training, we use that best lambda during the 3rd tier training for the regularization term corresponding to the first dataset.  For the lambda corresponding to the regularization term for the second dataset, we average the best lambdas for 2nd tier training permutations that started with that dataset, hypothesizing that this is a good approximation of optimal lambda for this term.  We then perform a small grid search around each of these values, testing the values 100 times less and 100 times greater than the estimated value and pick the best of these as optimal.  In picking the optimal lambdas for the 2nd and 3rd tiers of training, we choose the lambdas that have the least amount of drop in accuracy on the regularization term's corresponding dataset without too much drop in accuracy on the current training's dataset. We give optimal lambdas in Table \ref{tab:optimal_lambdas}.

\end{document}